\documentclass[conference]{IEEEtran}
\IEEEoverridecommandlockouts
\usepackage{cite}
\usepackage{amsmath,amssymb,amsfonts}
\usepackage{graphicx}
\usepackage{textcomp}
\usepackage{xcolor}
\usepackage{caption}
\usepackage{subcaption}
\usepackage{array}
\usepackage{multirow}
\usepackage{booktabs}

\def\BibTeX{{\rm B\kern-.05em{\sc i\kern-.025em b}\kern-.08em
    T\kern-.1667em\lower.7ex\hbox{E}\kern-.125emX}}
\begin{document}

\title{Improving Dietary Assessment Via \\ Integrated Hierarchy Food Classification
\thanks{$^{\star}$These authors contributed equally.}
}

\author{\IEEEauthorblockN{Runyu Mao$^{\star}$}
\IEEEauthorblockA{\textit{School of Electrical and Computer} \\
\textit{Engineering}\\
\textit{Purdue University}\\
West Lafayette, Indiana, USA \\
mao111@purdue.edu}
\and
\IEEEauthorblockN{Jiangpeng He$^{\star}$}
\IEEEauthorblockA{\textit{School of Electrical and Computer} \\
\textit{Engineering}\\
\textit{Purdue University}\\
West Lafayette, Indiana, USA \\
he416@purdue.edu}
\and
\IEEEauthorblockN{Luotao Lin}
\IEEEauthorblockA{\textit{Department of Nutrition Science} \\
\textit{Purdue University}\\
West Lafayette, Indiana, USA \\
lin1199@purdue.edu}
\and
\IEEEauthorblockN{Zeman Shao}
\IEEEauthorblockA{\textit{School of Electrical and Computer} \\
\textit{Engineering}\\
\textit{Purdue University}\\
West Lafayette, Indiana, USA \\
shao112@purdue.edu}
\and
\IEEEauthorblockN{Heather A. Eicher-Miller}
\IEEEauthorblockA{\textit{Department of Nutrition Science} \\
\textit{Purdue University}\\
West Lafayette, Indiana, USA \\
heicherm@purdue.edu}
\and
\IEEEauthorblockN{Fengqing Zhu}
\IEEEauthorblockA{\textit{School of Electrical and Computer} \\
\textit{Engineering}\\
\textit{Purdue University}\\
West Lafayette, Indiana, USA \\
zhu0@purdue.edu}
}
\maketitle
\IEEEpubidadjcol
\begin{abstract}
Image-based dietary assessment refers to the process of determining what someone eats and how much energy and nutrients are consumed from visual data. Food classification is the first and most crucial step. 
Existing methods focus on improving accuracy measured by the rate of correct classification based on visual information alone, which is very challenging due to the high complexity and inter-class similarity of foods. Further, accuracy in food classification is conceptual as description of a food can always be improved.
In this work, we introduce a new food classification framework to improve the quality of predictions by integrating the information from multiple domains while maintaining the classification accuracy. 
We apply a multi-task network based on a hierarchical structure that uses both visual and nutrition domain specific information to cluster similar foods. 
Our method is validated on the modified VIPER-FoodNet (VFN) food image dataset by including associated energy and nutrient information. 
We achieve comparable classification accuracy with existing methods that use visual information only, but with less error in terms of energy and nutrient values for the wrong predictions. 
\end{abstract}


\section{Introduction}
\label{sec:intro}
Assessing dietary intake accurately is challenging, yet an accurate profile of foods consumed is of paramount importance to reveal the true relationship of diet to health ~\cite{reedy2014higher}.  
Traditional dietary assessment is comprised of written and orally reported methods that can be time consuming and tedious and are not feasible for everyday monitoring \cite{poslusna2009misreporting,kirkpatrick2014performance}. Error is introduced primarily due to memory \cite{Thompson2017} and the human inability to accurately estimate food portion size \cite{schap2011}.

\begin{figure}[t]
\begin{center}
  \includegraphics[width=1.\linewidth]{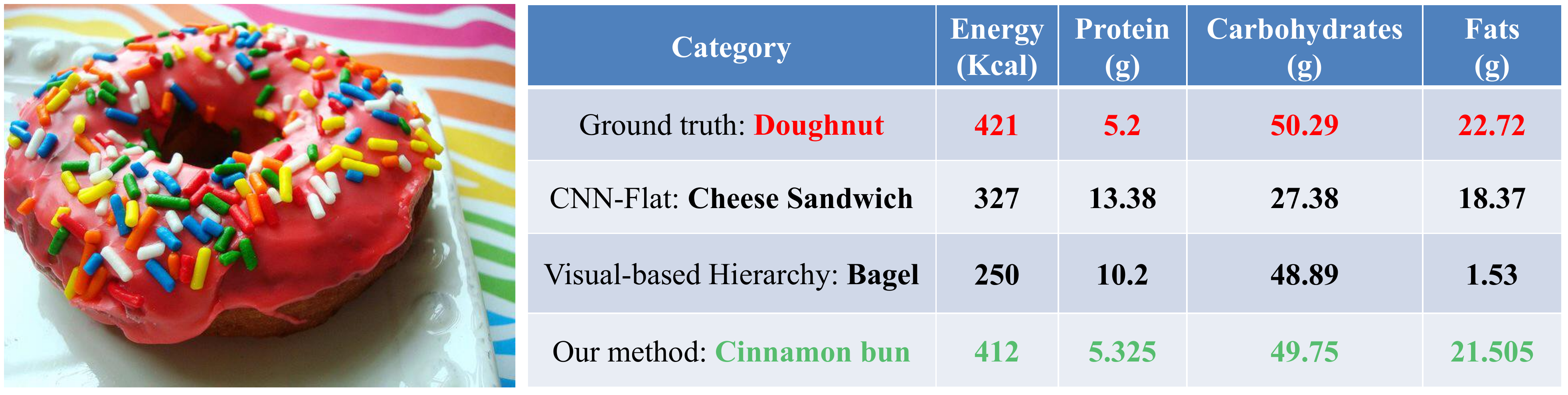}
  \caption{\textbf{An example of misclassification.} Comparison between flat CNN (non-hierarchical),  visual based hierarchy~\cite{mao2020visual} and our proposed method. Energy and nutrients ground truth are obtained from the USDA FNDDS database~\cite{fndds2018} using 100g food sample.}
  \label{fig:introduction}
\end{center}
\end{figure}

Recently, modern deep learning techniques~\cite{resnet,densenet,vgg} have enabled advances of image-based dietary assessment methods~\cite{mao2020visual,IBM,shao2021towards,yanai2015food,deepfood-liu2016,he2021end,foodnet-pandey2017,bolanos2016simultaneous,he2020multitask} which take only food or eating scene images as input and output the food types in the images. 
However, existing methods only focus on improving food classification accuracy measured by the rate of correctly classified foods based on visual information alone, which is very challenging due to the high complexity and inter-class similarity of foods.
Therefore, under the circumstance of noticeable classification error rate, a better mistake will enable a more accurate match of the food to a similar food within a food and nutrient database to improve the estimation of  energy or dietary components within the food compared with a ground truth. 
Figure~\ref{fig:introduction} shows an example of a misclassified food image where our proposed method gives the best prediction result in terms of energy and macronutrients values compared with flat (non-hierarchical) CNN and visual based hierarchy method~\cite{mao2020visual}.

The concept of hierarchy has been studied in deep learning based image classification~\cite{yan2015hd,2020bettermistakes}, which showed promising results. 
For food image classification, both~\cite{mao2020visual,IBM} proposed hierarchy based multi-task learning to improve the classification accuracy. However, a number of challenges remain to be addressed. 
Wu \textit{et al}.~\cite{IBM} proposed to use semantic hierarchy to learn the relationship between food categories. A better mistake is defined to have closer semantic meaning to the ground truth food category. However, the semantic hierarchical structure is manually generated for a specific food dataset~\cite{bossard14}, which cannot be generalized to other foods easily. In addition, food names are not unique for different geographic regions and are deeply rooted in local culture, making it challenging to build an adaptive semantic hierarchical structure. Recently, Mao \textit{et al}.~\cite{mao2020visual} proposed to build a visual hierarchical structure by extracting the visual features from CNNs and then measure the similarity score between food categories, which can be easily applied to different food image datasets. However, both methods achieved better mistakes measured by using visual information only, which is not sufficient for assessing dietary intake as illustrated in Figure~\ref{fig:introduction}. 

In this work, we introduce a novel cross-domain food classification framework to improve the quality of incorrect predictions in nutrition domain. We tackle both nutrition and visual domain information and embed them into a hierarchical structure for multi-task learning. Without losing the classification accuracy, our method is the first work to include nutrition domain information in food image classification.  Our method is evaluated on the modified version of the VIPER-FoodNet (VFN) food image dataset~\cite{mao2020visual}. The original VFN dataset contains more than 20k images of some of the most commonly consumed foods in the United States. We further enhanced this dataset by including the associated energy and macronutrients information (100g food sample) for each food category obtained from USDA Food and Nutrient Database for Dietary Studies (FNDDS) database~\cite{fndds2018}. 

\section{Our Proposed Method}
\label{sec:our method}
\begin{figure}[htp]
    \begin{center}
        \includegraphics[width=0.45\textwidth]{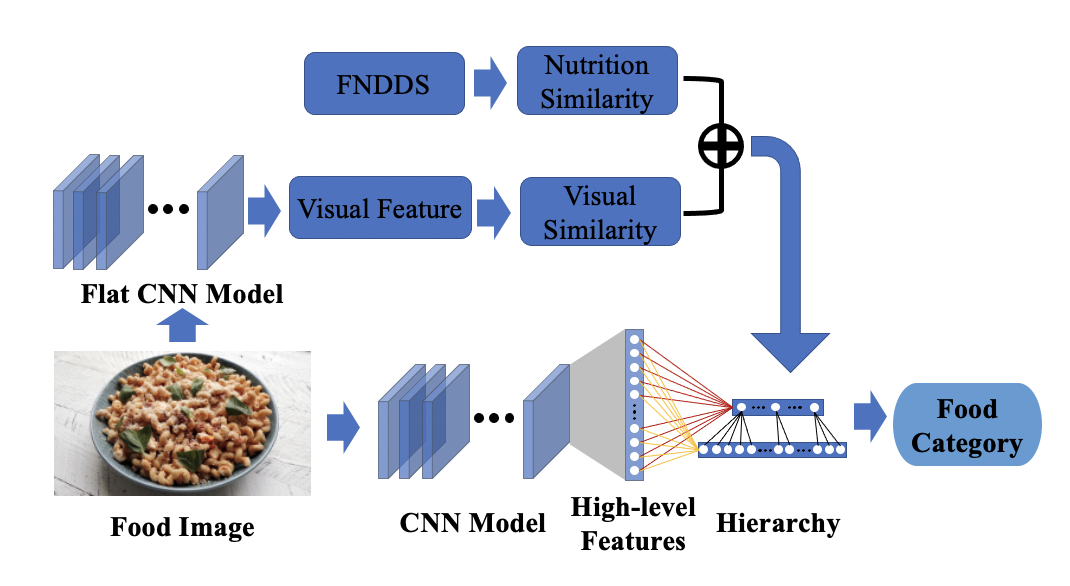}
        \caption{\textbf{Overview of our proposed food classification system.} A multi-task CNN model embeds the hierarchy that depicts the relations among food categories. The hierarchy is built based on nutrient information from the USDA FNDDS database and the visual features extracted by flat CNN model.}
      \label{fig:overview}
    \end{center}
\end{figure}

An overview of our proposed method is illustrated in Figure~\ref{fig:overview}. We first extract the visual feature map for each food image by training a flat CNN model using cross-entropy loss. Next, we calculate the similarity matrix based on both visual and nutrition domain information. Affinity Propagation (AP) is used to build a two-level hierarchy. Finally, we apply a multi-task network based on the obtained hierarchical structure for classification. 

\subsection{Cross-Domain Similarity Measure}
\label{ssec:similarity}

Systems of food classification have traditionally been defined based on the convention of how foods are included in a diet~\cite{wweia}. For example, there is not a scientific definition of a vegetable.
Each food is comprised of a panel of nutrients and the same nutrient may have significantly different distributions for different foods. In addition, dietary assessment often focuses on particular nutrients to evaluate the nutrition status of targeted population. For example, a low sodium diet and a fat-restricted diet weight the importance of sodium and fat differently. 
Accordingly, a good similarity measure based on nutrients should unify cross-domain nutrition information, and provide a fusion mechanism that can adapt to different dietary assessment needs.
In this work, we propose a similarity matrix to measure different nutrition information, \textit{e.g.}, energy, carbohydrate, fat and protein, and an adaptive parameter that can prioritize each contributing nutrient for a specific dietary assessment goal. 

Due to the wide range of nutritional values, the inter-class standard deviation is used as a scaling factor in the similarity measure as shown in Table~\ref{tab:interclass}.
We use the RBF kernel, Equation~\ref{eq:1}, to measure the similarity between different foods for each selected nutrition domain information, which is denoted by the nutrient value $x_{j}$, and the inter-class standard deviation $\sigma_{i}$. This results in a single similarity score $s_{i}$ normalized in the range (0,1] for each nutrient. We adopt the weighted harmonic mean, Equation~\ref{eq:2}, to dynamically unify $n$ different nutrition domain information for similarity measure. 
The similarity score of each nutrient is denoted as $s_{i}$ and weighted by parameter $w_{i}$. The final nutrient similarity score $S_{N}$ can be tuned for different dietary purposes by adjusting the parameter $w_i$.
\begin{equation} \label{eq:1}
   s_{i}(x_{1},x_{2}) = exp(-\frac{||x_{1}-x_{2}||^{2}}{2\sigma^{2}_{i}})
\end{equation}  
\begin{equation} \label{eq:2}
   S_{N}=\frac{\sum^{n}_{i=1}w_{i}}{\sum^{n}_{i=1}w_{i}s_{i}^{-1}} 
\end{equation}  
\begin{table}[t]
        \centering
        \caption{Selected Nutrients Inter-Class Standard Deviation}
        \begin{tabular}{|c|c||c|c|c|}
            \hline
            ~& Energy (kcal) & Carb. (g) & Fat (g) & Protein (g)\\
            \hline
             Std  & 119.48 & 16.96 &8.37 & 7.71\\
            \hline
            Range &[34.0, 595]&[0.71, 68.1]&[0.15, 51.7]&[0.16, 30.5]\\
            \hline
        \end{tabular}
        \label{tab:interclass}
\end{table}

To integrate the visual feature for similarity measure, we follow the procedure mentioned in~\cite{mao2020visual} which uses the output of the last feature extraction layer of the backbone network as the visual feature. 
We fit the histogram of each feature dimension to a Gaussian distribution, and calculate the overlapped coefficient (OVL)~\cite{OVL} to represent the similarity between food classes. Therefore, the visual similarity score, $S_{V}$ is also in the range (0,1]. We then normalize both the visual similarity matrix and nutrition similarity matrix to represent the overall inter-class similarity as the equally weighted harmonic mean:

\begin{equation} \label{eq:3}
   S=2\times\frac{S_{V}\times S_{N}}{S_{V}+S_{N}}
\end{equation}  
Therefore, the clustering result based on this similarity matrix could group food categories that are not only visually similar but also have similar nutrition values.

\subsection{Clustering and Hierarchical Multi-task Learning}
For food classification, as described in previous works~\cite{IBM,mao2020visual}, the hierarchical structure can be combined with a multi-task model to improve the performance by leveraging the inter-class relations embedded in the hierarchy. To build the hierarchical structure for our system, we needed to cluster similar food categories first. Based on the similarity matrix described in~\ref{ssec:similarity}, many clustering methods can be applied. In this paper, we choose the Affinity Propagation (AP) \cite{frey2007clustering} since it does not require a pre-defined number of clusters. We use AP to cluster similar food categories and generate a multi-level hierarchical structure. AP treats all food categories as candidates and selects $m$ candidates as exemplars to represent the $m$ clusters respectively. Such selection is iteratively refined until convergance.

Once the hierarchical structure is built, the multi-task model is employed for joint feature learning. In our implementation, we build a two level hierarchy where the upper-level nodes represent food clusters and the bottom level nodes represent food categories. Therefore, we have two tasks assigned to the multi-task model, \textit{i.e.}, $1^{st}$-level category predication and $2^{nd}$-level cluster prediction. The multi-task loss is formulated as:

\begin{equation} 
\label{eq:multi_loss}
L(\textbf{w}) = \sum_{t=1}^2\lambda_t\sum_{i=1}^{N_{t}} -log p(y_i^{(t)}|\textbf{x}_i,\textbf{w}_{0},\textbf{w}^{(t)}) 
\end{equation}
where $y_i^{(t)}$ is the corresponding class/cluster label for the $t^{th}$ hierarchical level. $\textbf{w}^{(t)}$ represent the network parameters for the $t^{th}$ output layer, and $\textbf{w}_{0}$ denote the parameters for the feature extraction layers. $\lambda_t$ is the hyperparameter that controls the weight of these two cross-entropy losses.
The joint feature learning will help the network extract the feature not only valuable for food category classification but also benefit cluster classification.

\section{Experimental Results}
\label{sec:exp results}
We select four nutrition information including Energy, Carbohydrate, Fat, and Protein calculated using 100g food samples and embed one or multiple of them in our food classification system. Since energy is not a nutrient, we do not combine it with other nutrient information in our classification system.
In all experiments, we fixed $w_{i}$ as 1 for nutrition similarity measure and set $\lambda_{t} = 1$ for multi-task learning.

\subsection{Dataset Preprocessing}
\label{prepare}
Since existing public food image datasets \cite{bossard14,wang2015recipe,Matsuda:2012ab,kawano2014automatic,cioccaJBHI,mixed_dish} do not contain information about the associated nutrient information for each food category, we build a unique dataset in this paper. We select the food images from the VIPER FoodNet (VFN) food image dataset~\cite{mao2020visual}, which contains 82 of the most frequently consumed foods in the U.S. based on What We Eat in America (WWEIA) food category classification~\cite{EM-nutrients2017}.
Each food category can map to multiple food codes and nutrient information in the Food and Nutrient Database for Dietary Studies (FNDDS) database~\cite{fndds2018}. 
Based on the FNDDS database, we made the following modifications to the VFN dataset. We combined `taco' and `tostada', `cake' and `cupcake' as they share the same nutrient information.  
Beverages were removed since they are typically assessed differently than foods.
A total of 74 food categories are selected for visual recognition. We collected 1,869 food items in FNDDS that belong to these food categories.  
Since a single food category contains multiple food items, we take the average of the nutrient values and energy information of all food items to represent the food category.

\subsection{Clustering Evaluation}
We evaluate the proposed clustering strategy based on the intra-cluster and inter-cluster relations from both nutrition and visual perspectives.  
Based on all the nutrition information we found in Section~\ref{prepare}, we calculate the intra-cluster and inter-cluster variances for each nutrition information,
\begin{equation} 
\label{eq:intra}
Var_{intra\text{-}cluster} = \frac{1}{N}\sum_{i}\sum_{j}(x_{ij}-\bar{X}_{i})^{2}
\end{equation}

\begin{equation} 
\label{eq:inter}
Var_{inter\text{-}cluster} = \frac{1}{N}\sum_{i}C_{i}(\bar{X}_{i}-\bar{X})^{2}
\end{equation}
where $N$ is the total number of images in the dataset. $C_i$ is the total number of images in the cluster $i$. $x_{ij}$ is the nutrition information of food category $j$ in cluster $i$. $\bar{X}_{i}$ is the mean value of cluster $i$, and $\bar{X}$ is the mean value of selected nutrition information on the entire dataset. 

Table~\ref{tab:inter_intra_var} summarizes the intra-cluster and inter-cluster variances of clustering results based on different nutrition information. Clustering solely based on the visual feature does not give good clustering results from nutrition perspective since the intra-cluster variance is very large and some are quite close to the inter-cluster variance. For other clustering results, we highlight the variances corresponding to the selected nutrition information. For example, ``F+P+V" means that the similarity measure for clustering considers fat, protein, and visual information and the results show that it has low intra-class variance and high inter-cluster variance for fat and protein. Note that it is not meaningful to compare the unselected nutrition information since it is not embedded in the clusters.

    \begin{table}[htbp]
        \centering
        \caption{Intra-cluster and inter-cluster variances of clustering results based on different nutrition information (Visual (V), Energy (E), Carbohydrate (C), Fat (F), Protein (P)).}
        \begin{tabular}{|c|c|c|c|c|c|c|}
            \hline
             ~ &  \multicolumn{6}{|c|}{$E (kcal^{2})$}\\
             \cline{2-7}
             ~&\multicolumn{3}{|c|}{intra $\downarrow$}&\multicolumn{3}{|c|}{inter $\uparrow$}\\
            \hline
            V&\multicolumn{3}{|c|}{9245.0}&\multicolumn{3}{|c|}{5029.5}\\
            E+V &\multicolumn{3}{|c|}{\textbf{1748.8}}&\multicolumn{3}{|c|}{\textbf{12525.7}}\\
            \hline
             ~ &  \multicolumn{2}{|c|}{$C (g^{2})$} & \multicolumn{2}{|c|}{$F (g^{2})$} &\multicolumn{2}{|c|}{$P (g^{2})$}\\
             \cline{2-7}
             ~&intra&inter&intra&inter&intra&inter\\
            \hline
            V & 187.8 & 99.9 & 56.5 & 13.5 & 28.3 & 31.1\\
            C+V & \textbf{16.4} & \textbf{271.3} & 55.5&14.5&26.7&32.7\\
            F+V & 163.6 & 124.1 & \textbf{6.2} &\textbf{63.8}&27.0&32.4\\
            P+V & 143.7 & 144.0 & 51.8 &18.2&\textbf{3.3}&\textbf{56.1}\\
            C+F+V & \textbf{77.5} & \textbf{210.2}&\textbf{8.6}&\textbf{61.3}&18.2&41.2\\
            C+P+V & \textbf{49.8} & \textbf{238.9} & 42.1&27.9&\textbf{3.7}&\textbf{55.7}\\
            F+P+V & 98.1 & 189.6 & \textbf{17.2}&\textbf{52.8}&\textbf{7.8}&\textbf{51.6}\\
            C+F+P+V & \textbf{53.0} & \textbf{234.7} & \textbf{22.3} &\textbf{47.7}&\textbf{8.6}&\textbf{50.8}\\
            \hline
        \end{tabular}
        \label{tab:inter_intra_var}
    \end{table}
   
To evaluate the visual similarity of the clustering results, we first generate a visual distance matrix $D_{V} =1 - S_{visual}$ where $S_{visual}$ is the visual similarity matrix containing similarity score $S_{V}$ in range (0,1]. The inter-cluster distance and intra-cluster distance are formulated as:

\begin{equation} 
\label{eq:intra_dist}
D_{intra\text{-}cluster} =\max_{i}(\frac{2}{ N_{i}(N_{i}-1)}\sum_{0<j<k<N_{i}}d_{jk})
\end{equation}

\begin{equation} 
\label{eq:inter_dist}
D_{inter\text{-}cluster} =\frac{2}{N_{c}(N_{c}-1)}\sum_{0<j<k<N_{c}}D_{jk}
\end{equation}
where $N_{i}$ is the total food category number in cluster $i$ and $d_{jk}$ is the visual distance between category $j$ and $k$ in cluster $i$. For intra-cluster distance, we choose the maximum distance, assuming worst case scenario. For inter-cluster distance, we choose the exemplar determined by Affinity Propagation (AP) as the centroid to represent the cluster. $D_{j,k}$ in Equation ~\ref{eq:inter_dist} is the distance between centroids of cluster $j$ and cluster $k$. $N_{c}$ is the total number of clusters. 
Table~\ref{tab:inter_intra_dist} summarizes visual distance of the intra-cluster, inter-cluster and the ratios between them for different clustering results. Clustering based on visual similarity alone has the best ratio. All other cases, which embed nutrition information, have higher ratio but still smaller than 1, indicating nutrition and visual information are successfully balanced and combined during clustering.  

    \begin{table}[t]
        \centering
        \caption{Visual distance of intra-cluster, inter-cluster and  ratio between them for different clustering results (Visual (V), Energy (E), Carbohydrate (C), Fat (F), Protein (P)).}
        \begin{tabular}{|>{\centering}p{0.08\textwidth}|>{\centering}p{0.08\textwidth}|>{\centering}p{0.08\textwidth}|>{\centering\arraybackslash}p{0.08\textwidth}|}
        \hline ~&\multicolumn{3}{c|}{Visual Distance}\\
       \cline{2-4}
       {} &intra $\downarrow$&inter $\uparrow$ &intra/inter $\downarrow$\\
       \hline
       V &0.4345&0.5302&0.8195\\
       E+V &0.4320&0.5026&0.8595\\
       C+V &0.4299&0.4803&0.8949\\
       F+V&0.4625&0.4673&0.9896\\
       P+V&0.4456&0.5222&0.8534\\
       C+F+V &0.4255&0.4995&0.8519\\
       C+P+V&0.4514&0.4913&0.9189\\
       F+P+V&0.4781&0.4993&0.9574\\
       C+F+P+V&0.4320&0.5072&0.8517\\
       \hline
        \end{tabular}
        \label{tab:inter_intra_dist}
    \end{table}

\subsection{Nutrition Analysis}
Since our proposed method is the first to consider cross-domain food classification by integrating nutrition-domain information, there is no previous methods that can be used for comparison. We implement two baselines for comparison in this section: (1) Flat training model: without considering the nutrient information and treat the food classification problem similar to image classification; (2) Visual based hierarchical classification~\cite{mao2020visual}: a previous work that considers visual similarity among foods to improve the prediction accuracy. 
For all methods, we use pre-trained ResNet-50~\cite{resnet} as the backbone classification network, and train our model with the Adam optimizer~\cite{adam}. The learning rate starts at 0.0001 and drops by a factor of 2 after every 5 epochs for a total of 50 epochs.
As shown in Table~\ref{tab:accuracy}, we found flat training classification and all hierarchy based classifications have comparable accuracy between $70 - 71\%$. This shows the integration of visual and nutrient information is well balanced and does not degrade the classification performance. 

    \begin{table}[t]
        \centering
        \caption{Classification accuracy for visual only flat training model, visual only hierarchy based model and integrated hierarchy based visual and nutrient domain model (Visual (V), Energy (E), Carbohydrate (C), Fat (F), Protein (P)).}
        \begin{tabular}{|c|c|c|c|c|c|}
       \hline
       ~ &flat&V&E+V&C+V&F+V\\
       \hline
       Accuracy & 70.13\%&70.33\%&70.31\%&70.68\%&70.54\%\\
       \hline
       ~ &P+V&C+F+V&C+P+V&F+P+V&C+F+P+V\\
       \hline
       Accuracy&70.91\%&70.84\%&70.70\%&70.73\%&70.70\%\\
       \hline
        \end{tabular}
        \label{tab:accuracy}
    \end{table}

\begin{figure}[htbp]
     \centering
     \begin{subfigure}[b]{0.23\textwidth}
         \centering
         \includegraphics[width=\textwidth]{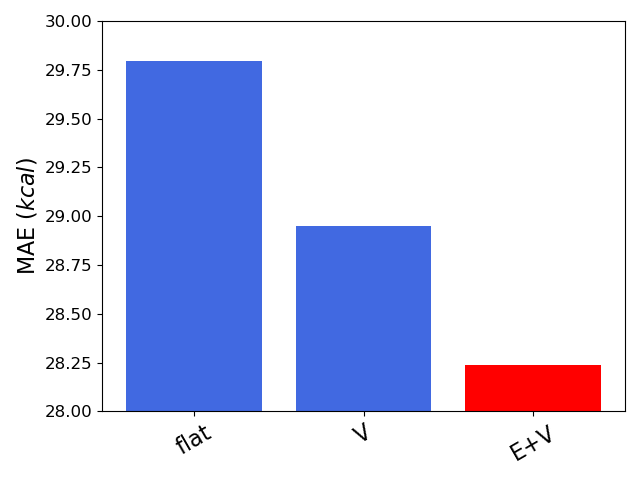}
         \caption{Energy Domain}
         \label{fig:E}
     \end{subfigure}
     \hfill
     \begin{subfigure}[b]{0.23\textwidth}
         \centering
         \includegraphics[width=\textwidth]{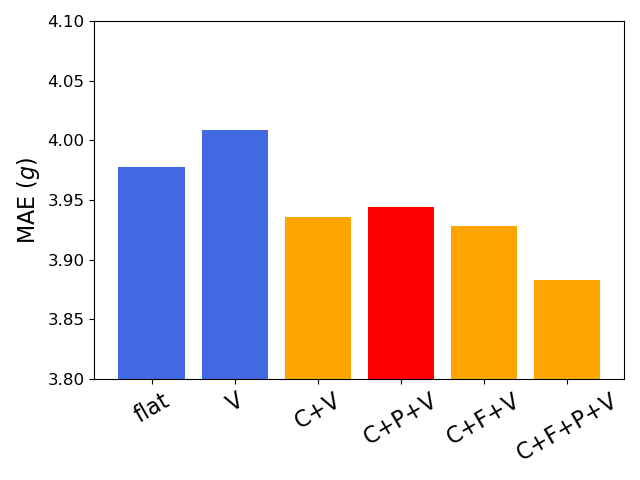}
         \caption{Carbohydrate Domain}
         \label{fig:C}
     \end{subfigure}
     \hfill
     \\
     \begin{subfigure}[b]{0.23\textwidth}
         \centering         \includegraphics[width=\textwidth]{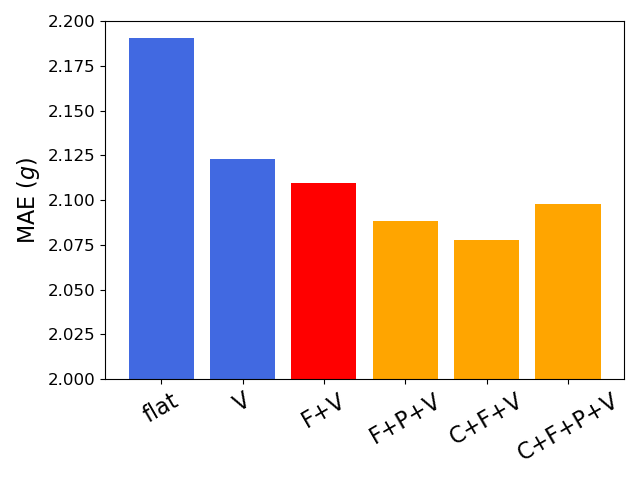}
         \caption{Fat Domain}
         \label{fig:F}
     \end{subfigure}
     \hfill
     \begin{subfigure}[b]{0.23\textwidth}
         \centering
         \includegraphics[width=\textwidth]{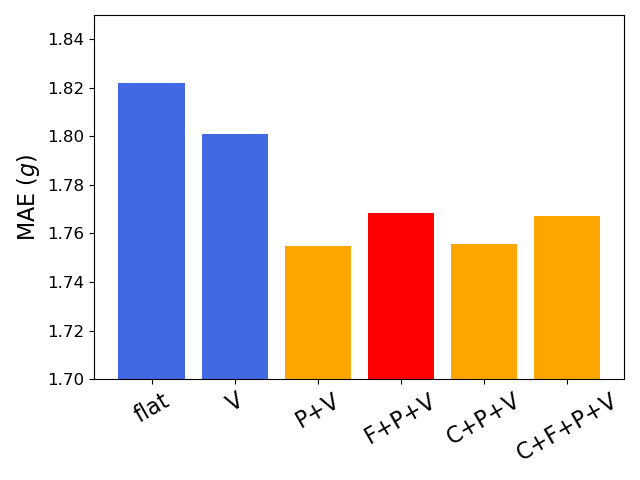}
         \caption{Protein Domain}
         \label{fig:P}
     \end{subfigure}
        \caption{Comparison of the Mean Absolute Error of nutrition information. The blue bars are classification methods without nutrition information embedded. Oranges bars represent our methods and red indicates the worst case scenario.}
        \label{fig:three graphs}
\end{figure}

In addition, we use the Mean Absolute Error (MAE) to evaluate whether our system makes better mistakes from nutrition perspective.
 \begin{equation} 
 \label{eq:mse}
 MAE = \frac{1}{N} \sum_{i=1}^{N} |(A_{i}-Y_{i})|
 \end{equation}
where $A_{i}$ is the nutrition value of the predicted food class, $Y_{i}$ is the nutrition value of the ground-truth category. 
The average results of hierarchy based food classification with nutrition information is shown in Figure~\ref{fig:E} - \ref{fig:P}. We treat the flat training and visual-based hierarchical classifications as baselines (blue bars) and compare with those that with nutrition information embedded (orange bars and red bar which is the worst case scenario).
Our nutrition embedded hierarchy classification method performs better compared with the baselines for all four nutrition information.

To better understand the improvement from nutrition domain information, we calculate the change in nutrition error compared to the flat training. 

 \begin{equation} 
 \label{eq:mse}
 relative~error~reduction = \frac{E_{i}-E_{flat}}{E_{flat}} 
 \end{equation}
 where $E_{i}$ is nutrient error of hierarchy based classification and $E_{flat}$ is the nutrient error of flat training model. 
Considering the worst case scenarios of our method, red bars in Figure~\ref{fig:three graphs}: energy+visual, carbohydrate+protein+visual, fat+visual, and fat+protein+visual, the proposed method still achieves 5.2\%, 0.8\%, 3.7\%, and 2.9\% relative error reduction, respectively. Although the three nutrients have similar range, carbohydrate is more challenging since its standard deviation is twice as large as shown in Table~\ref{tab:interclass}. All four methods using carbohydrate still achieves 1.4\% error reduction on average and the best case scenario shows 2.4\% improvement.

\section{Conclusion}
\label{sec:conclusion}
In this paper, we present a novel hierarchy based food classification framework to minimize prediction error by integrating both visual and nutrition domain information to improve the performance of food image classification. Our method is evaluated on an enhanced version of the VIPER FoodNet (VFN) food image dataset which includes associated energy and nutrient information for each food classes. We show that our method achieves similar classification accuracy compared with existing methods using visual information only, but minimized prediction error in terms of smaller MAE for the corresponding domain specific information, which shows great potential for assessing dietary intake with a focus on optimizing the accuracy of the nutrient panel information.

\bibliographystyle{IEEEtran}
\bibliography{ref}

\end{document}